\documentclass[conference]{IEEEtran}
\usepackage{cite}
\usepackage{graphicx}
\usepackage{xcolor}
\usepackage{booktabs}
\usepackage{url}
\usepackage{hyperref}
\usepackage{textcomp}
\usepackage{amsmath}

\begin{document}

\title{DriveNetBench: An Affordable and Configurable Single-Camera Benchmarking System for Autonomous Driving Networks}

\author{\IEEEauthorblockN{Ali Al-Bustami\IEEEauthorrefmark{1}\IEEEauthorrefmark{2}, Humberto Ruiz-Ochoa\IEEEauthorrefmark{1}, Jaerock Kwon\IEEEauthorrefmark{1}}
\IEEEauthorblockA{\IEEEauthorrefmark{1}Department of Electrical and Computer Engineering, University of Michigan-Dearborn, Dearborn, USA\\
Email: \{abustami, hruiz, jrkwon\}@umich.edu\\
\IEEEauthorrefmark{2}Corresponding Author}}
\maketitle

\begin{abstract}

Validating autonomous driving neural networks often demands expensive equipment and complex setups, limiting accessibility for researchers and educators. We introduce \textbf{DriveNetBench}, an affordable and configurable benchmarking system designed to evaluate autonomous driving networks using a single-camera setup. Leveraging low-cost, off-the-shelf hardware, and a flexible software stack, \textbf{DriveNetBench} enables easy integration of various driving models, such as object detection and lane following, while ensuring standardized evaluation in real-world scenarios. Our system replicates common driving conditions and provides consistent, repeatable metrics for comparing network performance. Through preliminary experiments with representative vision models, we illustrate how DriveNetBench effectively measures inference speed and accuracy within a controlled test environment. The key contributions of this work include its \textit{affordability}, its \textit{replicability} through open-source software, and its \textit{seamless integration} into existing workflows, making autonomous vehicle research more accessible. Code and pre-trained models are available at \href{https://github.com/alibustami/DriveNetBench}{\texttt{https://github.com/alibustami/DriveNetBench}}.

\end{abstract}

\begin{IEEEkeywords}
Autonomous Driving, Benchmarking, Single-Camera, Low-Cost, Replicability
\end{IEEEkeywords}

\section{Introduction}

Deep neural networks have become foundational to modern autonomous driving systems, enabling critical functions such as environment perception and decision-making processes. These encompass core tasks like object detection, path planning, and lane keeping. While multi-sensor configurations that incorporate LiDAR, radar, and multi-camera setups offer richer and more comprehensive data for training and testing, they necessitate costly hardware and complex integrations, increasing the entry barriers for both research and real-world deployment \cite{herman2017single}. Notably, research has shown that vision-based approaches can still achieve practical performance~\cite{bojarski2016end, coelho2022review, chib2023recent} highlighting the potential of camera-only systems to reduce costs and streamline experimentation.

Despite these advancements, standardized evaluation of autonomous driving networks remains a challenge \cite{kim2023opemi}. Many existing approaches rely on diverse simulation environments or large-scale datasets, making cross-network performance comparisons difficult. While simulations provide controlled conditions for algorithm development, they often fail to capture the unpredictable and dynamic factors that arise in physical, on-ground testing. Similarly, open-loop evaluations using pre-recorded datasets do not reflect how networks perform in closed-loop scenarios, where vehicle actions dynamically influence subsequent observations \cite{chen2024end}. 

\begin{figure}[!t]
    \centering
    \includegraphics[width=1\linewidth]{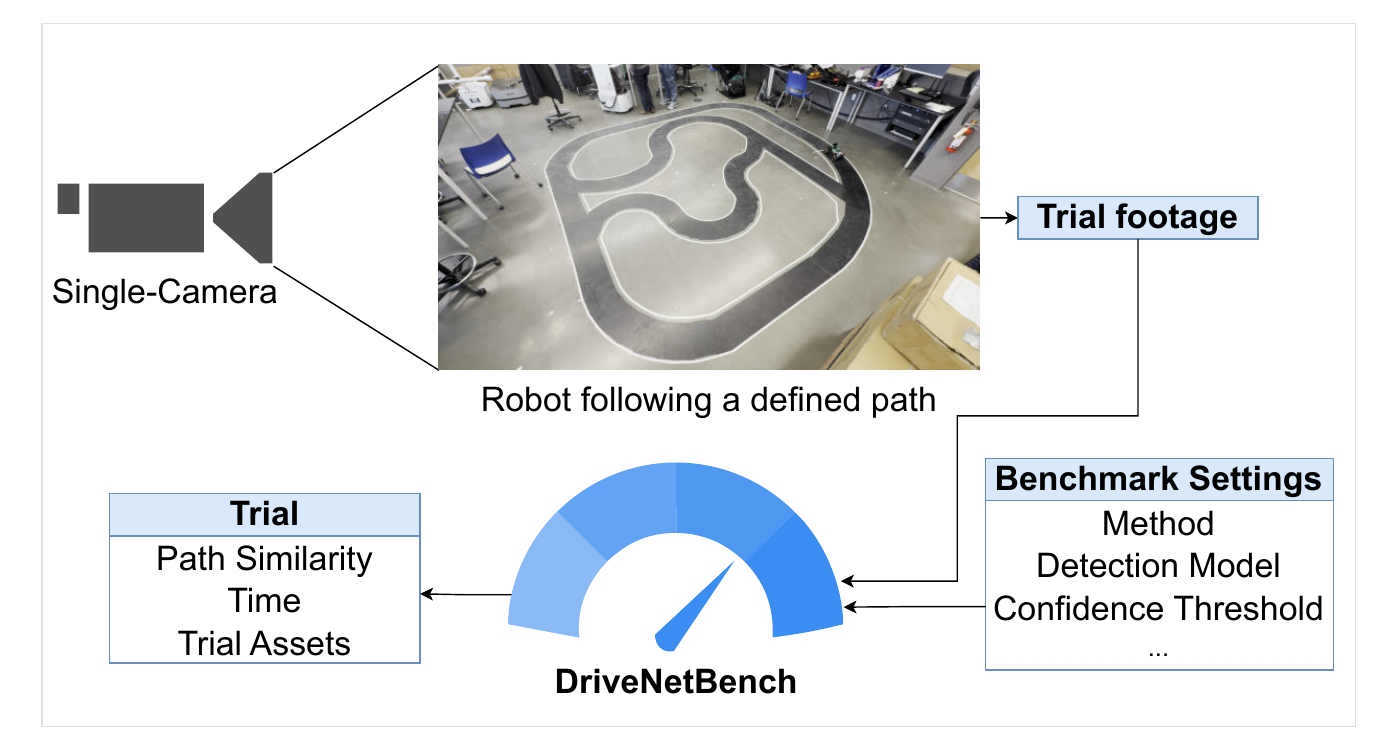}
    \caption{Overview of the DriveNetBench workflow: A single-camera captures robot trial footage, which DriveNetBench analyzes using configurable settings and driving performance metrics such as path similarity and completion time. The benchmark settings enable flexible evaluation of different models and parameters.}
    \label{fig:workflow}
\end{figure}

% Low-cost and scaled-down platforms \cite{boulet_guldner_karaman_2014, srinivasa2019mushr, paull2017duckietown, o2020f1tenth, balaji2020deepracer, khalil2021ridon} such as Duckietown~\cite{paull2017duckietown}, F1TENTH~\cite{o2020f1tenth}, and DeepRacer~\cite{balaji2020deepracer}, help mitigate some challenges in autonomous driving evaluation. However, they often lack standardized benchmarks tailored to fundamental driving tasks, such as lane following and obstacle avoidance, as well as repeatable testing conditions. On the other hand, high-fidelity simulation environments such as CARLA~\cite{dosovitskiy2017carla} provide robust testing capabilities but often require specialized hardware or expertise to configure. Recent benchmarking frameworks, including SafeBench~\cite{xu2022safebench} and DriveE2E~\cite{yu2024driveee}, have attempted to standardize evaluations but but primarily focus on complex multi-modal sensor setups or advanced simulation infrastructure. 

To address these challenges, we introduce \textbf{DriveNetBench}, a standardized benchmarking system designed to evaluate driving performance independently of the sensor suite used on the vehicle. DriveNetBench provides a unified evaluation framework for autonomous driving networks in a controlled environment, ensuring consistency across different setups. Figure~\ref{fig:workflow} illustrates the DriveNetBench workflow: a single-camera captures robot trial footage, which is analyzed using configurable settings and driving performance metrics such as path similarity and completion time.
DriveNetBench is designed to be affordable, modular, and highly configurable, enabling accessible and standardized evaluation of autonomous driving networks. Specifically, our contributions include:

\begin{enumerate}
\item \textbf{Affordability and Easy Replication:} We outline the design of a compact hardware kit built from readily available consumer-grade parts, enabling researchers, educators, and hobbyists to replicate the setup at minimal cost.
\item \textbf{Modular Benchmarking Pipeline:} Our open-source software framework provides a standardized interface for capturing data, integrating driving networks, and computing standard evaluation metrics.
\item \textbf{Configurable Experimentation:} Users can easily adjust benchmark parameters such as track layout, threshold definitions, and evaluation metrics, via simple configuration files, eliminating the need for code modifications.
\item \textbf{Comprehensive Metric Analysis:} DriveNetBench reports both accuracy-based metrics (e.g., path similarity) and efficiency-based metrics (e.g., completion time), offering a holistic assessment of driving performance.
\end{enumerate}

We demonstrate the effectiveness of DriveNetBench by integrating multiple driving models, and comparing their closed-loop performance on the same track under identical conditions. Our findings highlight how standardized metrics reveal trade-offs among different approaches, promoting fair comparisons and accelerating iterative development. We envision DriveNetBench as a valuable resource for researchers seeking cost-effective yet rigorous real-world validation, as well as for educators aiming to teach autonomous driving concepts through hands-on experimentation.

\section{Related Work}

Many benchmarking efforts in machine learning and autonomous systems focus on peak algorithmic accuracy or performance on curated datasets, often overlooking practical constraints like hardware costs or real-world deployment challenges. For instance, MLPerf~\cite{mattson2020mlperf} has established itself as a standard for measuring machine learning inference and training throughput, yet it can necessitate specialized compute clusters, limiting adoption in smaller labs. Similarly, EEMBC’s CoreMark~\cite{coremark} is valued for its simplicity and low overhead but is geared towards benchmarking core CPU performance, leaving out the holistic needs of an autonomous driving pipeline.

Low-cost and scaled-down platforms \cite{boulet_guldner_karaman_2014, srinivasa2019mushr, paull2017duckietown, o2020f1tenth, balaji2020deepracer, khalil2021ridon} such as Duckietown~\cite{paull2017duckietown}, F1TENTH~\cite{o2020f1tenth}, and DeepRacer~\cite{balaji2020deepracer}, help mitigate some challenges in autonomous driving evaluation. However, they often lack standardized benchmarks tailored to fundamental driving tasks, such as lane following and obstacle avoidance, as well as repeatable testing conditions. On the other hand, high-fidelity simulation environments such as CARLA~\cite{dosovitskiy2017carla} provide robust testing capabilities but often require specialized hardware or expertise to configure. Recent benchmarking frameworks, including SafeBench~\cite{xu2022safebench} and DriveE2E~\cite{yu2024driveee}, have attempted to standardize evaluations but primarily focus on complex multi-modal sensor setups or advanced simulation infrastructure. 

Our proposed benchmark, DriveNetBench, is designed to be minimalistic and easily replicable, making it suitable for both research and educational applications. Unlike existing benchmarks, DriveNetBench is independent of the sensor suite mounted on the vehicle and does not rely on a specific vehicle software stack, allowing it to be seamlessly adapted to various small-scale robots or vehicles. For instance, DriveNetBench can be used to directly compare the driving performance of an F1TENTH\cite{o2020f1tenth} car against a DeepRacer\cite{balaji2020deepracer} car under identical conditions. Furthermore, DriveNetBench standardizes evaluation metrics such as path similarity and completion time, ensuring consistent and reproducible assessments across diverse driving models. 

By providing an accessible, transparent, and standardized evaluation framework, DriveNetBench aligns with broader open-science objectives~\cite{stodden2014implementing}, promoting reproducible and transparent experimentation. Designed to be low-cost and easily replicable, DriveNetBench significantly reduces the technical, financial, and logistical barriers to real-world experimentation while maintaining consistent performance evaluation.

\section{DriveNetBench System Overview}
\label{sec:system_overview}
\begin{figure}[t]
\centering
\includegraphics[width=1\columnwidth]{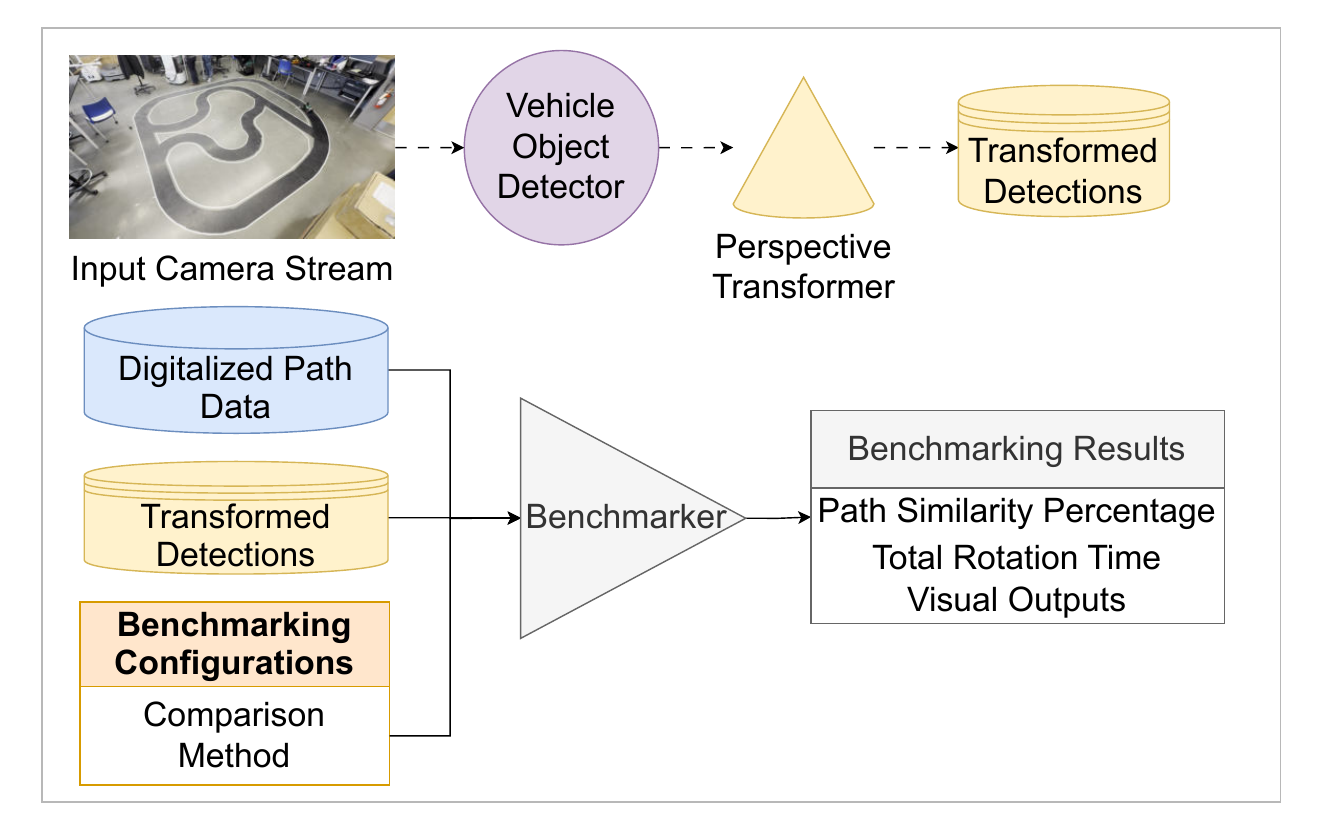}
\caption{Overview of the DriveNetBench system architecture. The system detects vehicles from a camera feed, applies a perspective transform, and compares against digitalized path data, outputting path similarity, rotation time, and visual diagnostics.}
\label{fig:arch}
\end{figure}

DriveNetBench is designed as a modular system mirroring a typical autonomous driving pipeline for standardized evaluation. Figure~\ref{fig:arch} illustrates the overall architecture. The workflow is composed of four main elements:

\begin{enumerate}
    \item \textbf{Input Data Source:} A single overhead (or side-mounted) camera records a physical track where a mobile robot or small-scale vehicle navigates. 
    \item \textbf{Network Under Test and Transformation Module:} The video stream or saved trial footage is fed to the ``Network Under Test'' (treated as a black box), and then to a \emph{Transformation Module} that aligns detections with the coordinate system of a ``digital twin'' of the track.
    \item \textbf{Evaluation and Metrics Module:} The system computes quantitative metrics (e.g., path similarity, completion time) by comparing the robot’s driven trajectory against a ground-truth path.
    \item \textbf{Supporting Modules:} Auxiliary components (Figure~\ref{fig:supporting_modules}) automate tasks like track definition, keypoint-based homography calibration, and robot detection.
\end{enumerate}

\subsection{Supporting Modules and Track Definition}
\begin{figure*}[t]
    \centering
    \includegraphics[width=0.95\textwidth]{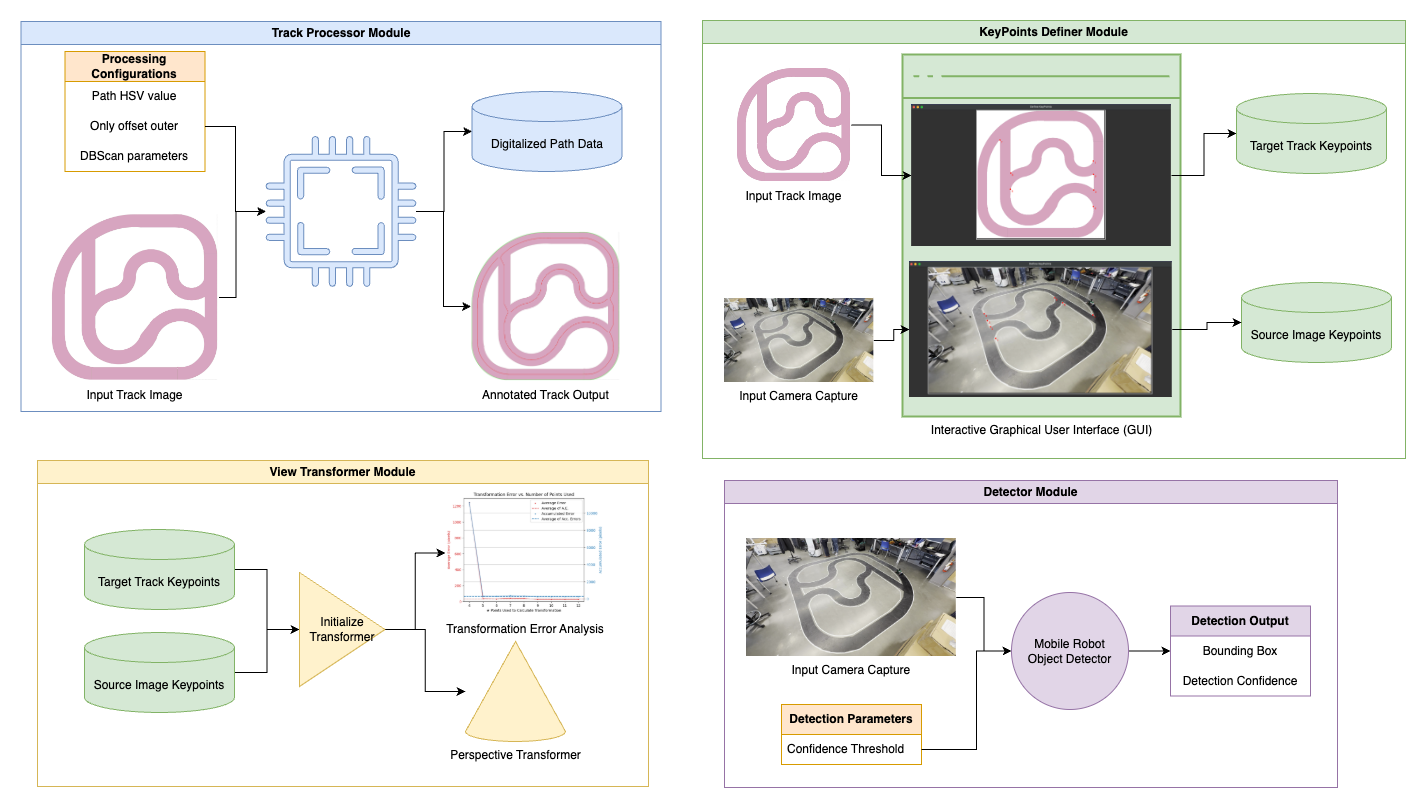}
    \caption{Supporting modules of DriveNetBench, demonstrating how the 
    \emph{TrackProcessor}, \emph{KeyPoints Definer}, \emph{View Transformer}, 
    and \emph{Detector} modules function together to facilitate path generation, 
    keypoint alignment, perspective transformations, and robot detection.}
    \label{fig:supporting_modules}
\end{figure*}

As shown in Figure~\ref{fig:supporting_modules}, DriveNetBench relies on several supplementary modules to ensure consistent experiment setup. 

\begin{itemize}
    \item \textbf{TrackProcessor:} Given a digital representation of the track (a ``digital twin''), the system extracts the centerline or pre-defined route using image processing. This yields the target reference path.
    \item \textbf{KeyPoints Definer:} Ensures accurate homography estimation by defining pixel coordinates in the recorded footage and the corresponding points in the digital track space. 
    \item \textbf{View Transformer:} Uses the calculated homography to project real-time detections from the camera frame to the digital twin’s coordinate system.
    \item \textbf{Detector:} A YOLO-based~\cite{Jocher_Ultralytics_YOLO_2023} (or other) model that locates the robot in each frame. We then extract the robot’s centroid for trajectory analysis.
\end{itemize}

By handling these tasks in modular fashion, DriveNetBench provides flexibility to accommodate different cameras, track designs, and detection models.

\subsection{Network Under Test}
While the \emph{Supporting Modules} handle track-related tasks, the \textbf{Network Under Test} represents the autonomous driving model under evaluation. Because DriveNetBench treats this network as a black box, researchers can integrate any vision-based solution with minimal friction. The system ensures that each network is evaluated under identical conditions, providing fair comparisons even if the models differ in architecture or hardware requirements.

\subsection{Evaluation and Metrics}
DriveNetBench records the robot’s driven trajectory over time and compares it to the ground-truth reference path (Section~\ref{sec:approach_metrics} details the metrics). Key measures include:

\begin{itemize}
    \item \textbf{Path Similarity:} How well the driven path aligns with the reference route.
    \item \textbf{Completion Time:} The duration required to complete a route or circuit. Failures (e.g., leaving the track) can be penalized appropriately.
\end{itemize}

Additionally, the system tracks homography-related errors (transformation accuracy) to ensure that any miscalibration is caught before final scoring. 

\section{Benchmarking Approach and Metrics}
\label{sec:approach_metrics}
DriveNetBench provides standardized scenarios and collects metrics that capture both performance and reliability.

\subsection{Path Similarity}
We compute how closely the driven path approximates a reference route using distance-based metrics between two sets of 2D points:
\[
\mathbf{x} = \{x_1, x_2, \ldots, x_{n}\}, 
\quad
\mathbf{y} = \{y_1, y_2, \ldots, y_{m}\}.
\]
We offer two distance metrics:

\begin{itemize}
\item \textbf{Dynamic Time Warping (DTW):} Aligns path indices by allowing non-linear time shifts.
\item \textbf{Fr\'echet Distance:} Measures the minimum leash length for simultaneously traversing both paths.
\end{itemize}

An optional \emph{clamping} threshold $\delta$ can bound extreme deviations. After computing the distance $d(\mathbf{x}, \mathbf{y})$, DriveNetBench normalizes it by a baseline $B$, producing a final similarity score $S \in [0\%,100\%]$:

\begin{equation}
S = \min\!\Bigl(100\%,\ 
100\% \times \max\Bigl(0,\ 1 \;-\; \frac{d(\mathbf{x},\mathbf{y})}{B}\Bigr)\Bigr).
\end{equation}

\subsection{Completion Time}
Completion time is the total time spent driving until the route or circuit is finished. If a model fails to stay on track or to reach the endpoint, we record a failure event or large penalty. By comparing these metrics alongside path similarity, DriveNetBench elucidates trade-offs between precise lane-following and quick traversal.

\subsection{Transformation Error Analysis}
Because DriveNetBench projects robot detections to the digital track, any homography miscalibration can inflate path errors. We therefore log both \emph{average error} and \emph{accumulated error} by re-checking the alignment of known calibration points. If error values exceed a threshold, users can refine or add keypoints.

\begin{figure}[t]
    \centering
    \includegraphics[width=0.8\columnwidth]{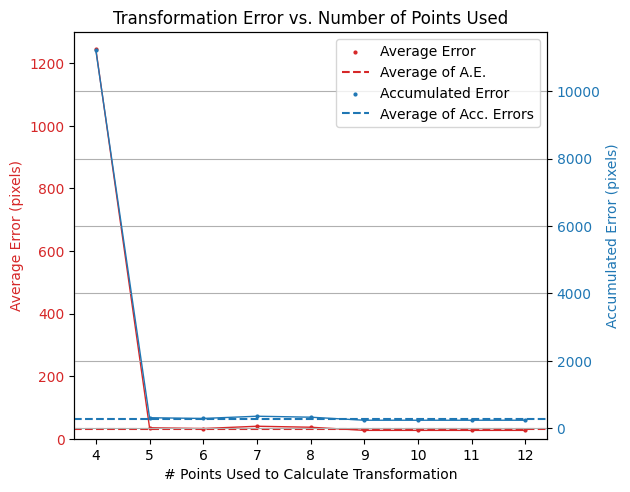}
    \caption{Transformation error vs.\ number of homography keypoints. Accumulated error (blue) and average error (red) sharply decrease once five or more points are used.}
    \label{fig:transform_error}
\end{figure}

Figure~\ref{fig:transform_error} shows how these errors decrease sharply after using at least five keypoints in the homography estimation, flattening out once the transformation is well-constrained.

\section{Experimental Results}
\label{sec:results}
\begin{figure}[t]
    \centering
    \includegraphics[width=0.6\columnwidth]{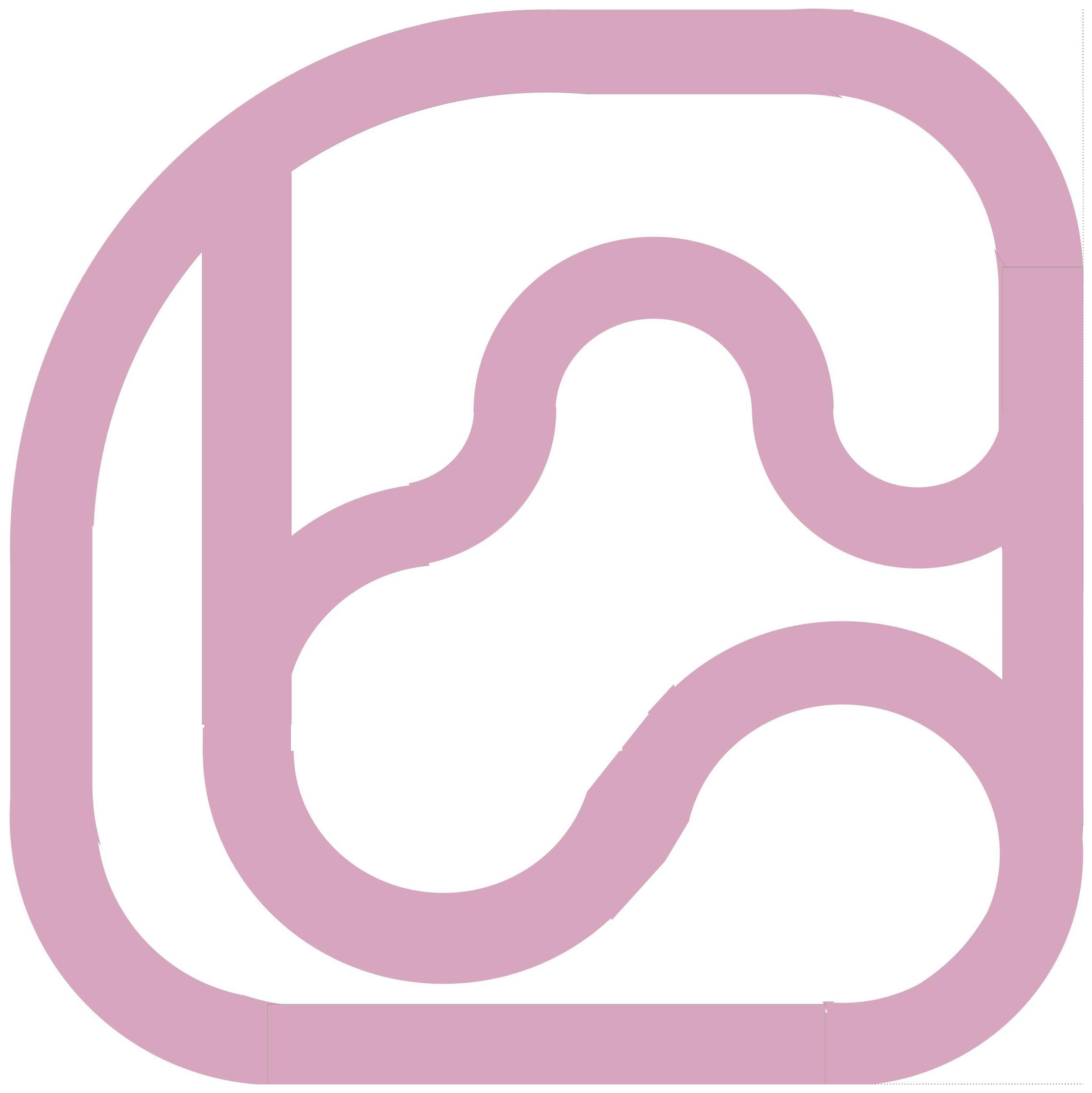}
    \caption{Example digital track (``digital twin'') from which a reference route is extracted.}
    \label{fig:digital_track}
\end{figure}

We validated DriveNetBench in an indoor lab using a single overhead camera and a small autonomous robot navigating the track from Figure~\ref{fig:digital_track}. The camera was mounted at a height of approximately 2.15\,m with a pitch angle of about 41\textdegree. Although an OSCAR BIMINet network~\cite{kwon_oscar_2021, kwon2022incremental} was integrated for lane-following, the specific trials reported here use human-driven data (four different drivers), solely for illustrative purposes. 

Table~\ref{tab:oscar_perf} highlights sample outcomes from four drivers. Path similarity scores show how closely each driver maintained the centerline, while completion time reflects the duration. These results confirm that DriveNetBench can consistently capture fundamental driving metrics under uniform conditions. In a similar manner, any neural driving model can be swapped in with minimal code changes, enabling standardized benchmarking of algorithmic performance.

\begin{table}[t]
    \caption{Illustrative performance logs for four different (human) drivers on the same track.}
    \label{tab:oscar_perf}
    \centering
    \begin{tabular}{lcc}
    \toprule
    \textbf{Driver} & \textbf{Path Sim. (\%)} & \textbf{Completion Time (s)} \\
    \midrule
    1 & 95.78 & 67.5 \\
    2 & 92.68 & 53.4 \\
    3 & 88.62 & 42.9 \\
    4 & 84.57 & 34.7 \\
    \bottomrule
    \end{tabular}
\end{table}

\section{Conclusion}
We have presented \textbf{DriveNetBench}, an affordable and configurable single-camera system for benchmarking autonomous driving networks. By adopting a minimal hardware design and an open-source, modular software pipeline, DriveNetBench lowers the cost and complexity of real-world evaluations. Key advantages include:

\begin{itemize}
\item \emph{Affordability \& Replicability:} Off-the-shelf components and a concise calibration procedure allow quick, low-cost setup.
\item \emph{Flexible Experimentation:} Parameterized benchmark configurations enable a wide range of driving tasks, from lane-following to obstacle avoidance.
\item \emph{Standardized Metrics:} Uniform scoring of path similarity and completion time ensures fair comparisons, while transformation error logs help diagnose calibration issues.
\end{itemize}

Looking ahead, we plan to extend DriveNetBench to support stereo cameras, additional sensors, and larger-scale outdoor tracks. By open-sourcing our hardware designs and software code, we aim to foster a collaborative community where researchers can easily compare models, share improvements, and accelerate development in autonomous driving. DriveNetBench stands poised to fill a critical gap by providing a low-cost yet robust testbed that bridges the space between purely simulated environments and expensive full-scale test vehicles.

\subsection{Limitation and Future Works}

While our single-camera focus offers significant advantages in affordability and accessibility, it also poses certain limitations compared to multi-sensor systems (e.g., LiDAR or radar). First, single-camera setups do not provide direct depth information, making real-time distance estimation more challenging under varied lighting or weather conditions. Second, the reliance on purely monocular imagery is more sensitive to occlusions, lens distortions, and calibration drift, which can impact detection accuracy. Despite these shortcomings, our goal with DriveNetBench is to lower the financial and technical barriers to real-world testing, enabling researchers and educators to iterate quickly on algorithmic improvements. Looking ahead, we plan to explore optional modules for depth estimation or multi-sensor fusion, thus extending the current single-camera paradigm to capture a broader range of autonomous driving scenarios.
\bibliographystyle{IEEEtran}
\bibliography{references.bib}

\end{document}